\documentclass[11pt]{article}

\usepackage{placeins}
\usepackage{amsfonts}
\usepackage{amsmath}
\usepackage{tcolorbox }

\usepackage[preprint]{acl}

\usepackage{times}
\usepackage{latexsym}

\usepackage[T1]{fontenc}

\usepackage[utf8]{inputenc}

\usepackage{microtype}

\usepackage{inconsolata}

\usepackage{graphicx}

\title{Improving reasoning at inference time via uncertainty minimisation}

\author{Nicolas Legrand \\
  Center for Humanities Computing \\
  Aarhus University, Denmark \\
  \texttt{nicolas.legrand@cas.au.dk}
  \And
  Kenneth Enevoldsen \\
  Center for Humanities Computing \\
  Aarhus University, Denmark \\
  \texttt{kenneth.enevoldsen@cas.au.dk} 
  \AND
  Márton Kardos \\
  Center for Humanities Computing \\
  Aarhus University, Denmark \\
  \texttt{martonkardos@cas.au.dk} 
  \And
  Kristoffer Nielbo \\
  Center for Humanities Computing \\
  Aarhus University, Denmark \\
  \texttt{kln@cas.au.dk}
  }

\begin{document}
\maketitle
\begin{abstract}
Large language models (LLMs) now exhibit strong multi-step reasoning abilities, but existing inference-time scaling methods remain computationally expensive, often relying on extensive sampling or external evaluators. We propose a principled strategy that frames reasoning as uncertainty minimisation and operates at the level of individual thoughts rather than tokens. Our method selects, at each reasoning step, the continuation that maximizes the model's self-certainty, a metric computed from its internal predictive distribution. This approach achieves significant improvement with a small number of samples, relies exclusively on model-internal signals, and applies to open-ended questions as opposed to methods like majority voting. Experiments on MATH500 and GSM8K across multiple model sizes demonstrate that thought-level self-certainty maximization consistently outperforms greedy decoding and matches or exceeds self-consistency under comparable token budgets. Cross-linguistic evaluations further indicate that the method transfers robustly beyond high-resource languages. Furthermore, analysis of self-certainty dynamics reveals that correct reasoning trajectories converge early to stable paths, suggesting that early decisions, likely associated with the planning of the reasoning process, are predictive of final accuracy. Building on this result, we show that self-certainty maximisation applied to the early steps can explain most of the performance gain and provide a simple yet efficient inference-time scaling method.
\end{abstract}

\begin{figure*}[htbp]
\includegraphics[width=\linewidth]{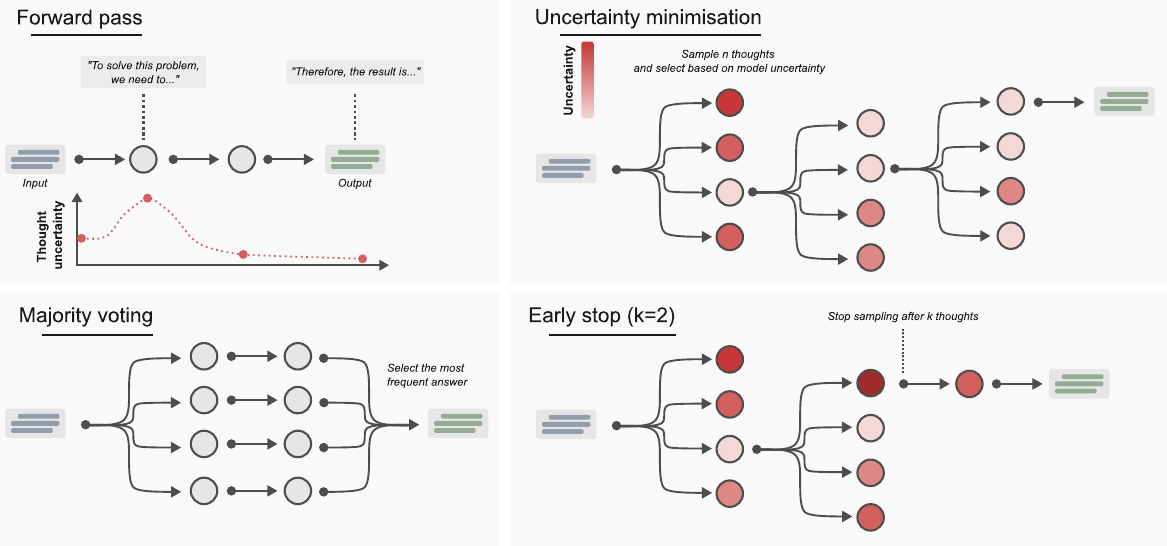}
\caption{\small \textbf{Inference time scaling using uncertainty minimisation.} We posit that reasoning in LLMs and humans proceeds by uncertainty minimisation. Each reasoning step, is selected for its potential to reduce uncertainties towards a required solution. We emulate such a selection process by sampling multiple candidates at every step and retaining the generation with the highest self-certainty score.
}
\label{fig:design}
\end{figure*}

\section{Introduction}

Large Language Models (LLMs) now exhibit strong performance on complex multi-step reasoning tasks \citep{Xu:2025, Shao:2024, deepseekr1}, driven by prompting strategies that decompose problems into sequences of intermediate steps, such as Chain-of-Thought prompting (CoT) \citep{Wei:2022}, and tree-based extensions \citep{Zhou:2023}.

Recent work has therefore focused on inference-time scaling, allocating additional computation during generation rather than training \citep{zhang:2025}. This has proven highly effective for mathematics and programming \citep{openaio1}, enabling smaller models to approach the performance of most efficient systems \citep{beeching:2024}. Methods range from repeated sampling and self-consistency or best-of-\textit{n} selection \citep{Brown:2024, Wang:2022}; beam search guided by process reward models (PRMs) \citep{Zhou:2023}; and more recently, particle-based or probabilistic search schemes \citep{Puri:2025}. However, they all operate either at the token level, where local uncertainty can be noisy and misleading, or at the full generation level, thereby ignoring the dynamic structure of reasoning and requiring expensive rollouts.

Similarly, evaluating reasoning steps can be costly if performed by an external model. Therefore, methods have explored the use of models' internal signals to guide generation instead. Among these signals, uncertainty minimization has shown particularly strong results, with internally confident generations often correlating with higher reasoning accuracy \citep{Agarwal:2025, Zhao:2025}. But these methods operate either on fully generated chains or at the token level. Neither granularity aligns well with how reasoning is thought to unfolds at a cognitive level \citep{Kargupta:2025}, for example, as a sequence of semantically coherent intermediate steps progressing toward uncertainty resolution.

In this paper, we argue that the appropriate unit of analysis for uncertainty-driven reasoning is the 'thought level': intermediate steps produced during CoT generation. Thought-level signals capture conceptual subunits of reasoning, control for variable generation length, and preserve the temporal structure of confidence evolution. We propose an inference-time method that selects reasoning steps by maximizing self-certainty, guiding generation toward paths that consistently reduce uncertainty. 

Crucially, reasoning is a dynamic process: uncertainty may transiently increase before collapsing as a coherent plan emerges. An important component in this regard is meta-cognitive control strategies \citep{Kargupta:2025}, such as goal management \citep{Griffiths2019} and strategy selection \citep{Lieder2017}. Similar steps are likely to take place early in LLMs as well which could provide further guidance for focusing inference budget optimally. 

Finally, the mechanistic connection between thoughts is unclear in LLMs \citep{Lanham:2023,Wang2025,Zolkowski:2025}, which could compromise the effectiveness of uncertainty minimisation on edge cases. As a robustness check, we further evaluate the effect of this method on a low-mid resources language Danish.

In summary, our key contributions are as follows:

\begin{enumerate}
  \item We introduce a new inference-time scaling method, extending uncertainty-based signals to the granularity of individual reasoning steps \citep{Kang:2025, Zhao:2025}.
  \item We evaluate its scaling behaviour on MATH500 \citep{Lightman:2023} and GSM8K across multiple model sizes in the Qwen \citep{qwen:2024} and Llama \citep{llama} families.
  \item We identify characteristic temporal patterns of self-certainty that predict downstream reasoning correctness, offering insight into internal dynamics of LLM reasoning and optimisation principles for inference scaling.
  \item We investigate cross-linguistic generalisation by evaluating our method on Danish translations of GSM8K \citep{gsm8k}, assessing whether reasoning strategies transfer across typologically different languages.
\end{enumerate}

\FloatBarrier
\section{Related Work}

\textbf{Inference time scaling.} Inference-time scaling refers to methods that improve LLM performance by allocating additional computation during generation rather than modifying model parameters \citep{zhang:2025}. This paradigm has proven particularly effective for reasoning-intensive tasks such as mathematics and programming \citep{openaio1, beeching:2024}, allowing smaller models to approach the performance of much larger systems. Existing approaches differ in their reliance on external evaluators, compute allocation strategies, and decision granularity. Methods relying on external scorers, such as beam search guided by process reward models \citep{Zhou:2023,Snell:2024}, or Monte Carlo Tree Search (MCTS; \citealp{Guan:2025}), are effective but introduce additional training requirements. Verifier-free methods rely exclusively on the base model's outputs. Examples include repeated sampling strategies such as Best-of-N (BoN; \citealp{Cobbe:2021}) and self-consistency with majority-voting \citet{Wang:2022}, which select answers based on frequency or final-score aggregation. These methods require many rollouts, sometimes hundreds per query \citep{Brown:2024}. For that reason, recent work tried to improve efficiency with strategic allocation of sampling budget, for example, through adaptive sampling based on uncertainty \citet{Huang:2025} or distance to solution \citep{Chatziveroglou:2025}, or by exploring the typical set of trajectories via particle-based methods \citet{Puri:2025}. Most of these methods, however, work either at the token level or at the level of full generation (but see \citealp{Chatziveroglou:2025} for an example of step-level analysis).

\textbf{Uncertainty estimation in LLMs} LLMs are known to be miscalibrated and to overestimate confidence \citet{Xiong:2024,Yang:2024}, contributing to fluent but incorrect outputs \cite{Fadeeva:2023} and limiting reliable metacognitive assessment \citep{Griot2025}. Token-level uncertainty metrics, using the predicted log-probabilities in the final token embedding, are difficult to interpret as they conflate epistemic and aleatoric uncertainty. Several approaches estimate uncertainty at higher levels of abstraction, including semantic entropy, which clusters multiple generations to detect hallucinations or incorrect answers \citep{Farquhar2024}. While effective for post hoc detection, such methods require multiple full generations and are poorly suited for online control. Complementing these approaches, \citet{Kang:2025} introduced self-certainty, defined as the average Kullback–Leibler divergence between the predicted token distribution and a uniform distribution. This measure avoids several shortcomings of entropy-based metrics, which can be biased toward longer or superficially confident generations, and was shown to be a robust predictor of final accuracy.

\textbf{Uncertainty minimisation as a control signal} Uncertainty minimisation at inference time assumes that well-trained models should exhibit increasing correctness likelihood as their internal confidence grows. This principle has been applied both in post-training and during inference. \citet{Agarwal:2025} showed that entropy minimisation can improve reasoning at the token level. \citet{Zhao:2025} leveraged internal feedback via Group Relative Policy Optimisation (GRPO) to train models to reason more effectively. \citet{wang:2025} have used the uncertainty of the previous generation as a signal to guide the allocation of compute using a multi-armed bandit framework. \citealt{Huang:2025} have used an uncertainty-based measure to stop the generation of confident responses and avoid unnecessary token usage.

\begin{figure*}[htbp]
\includegraphics[width=\linewidth]{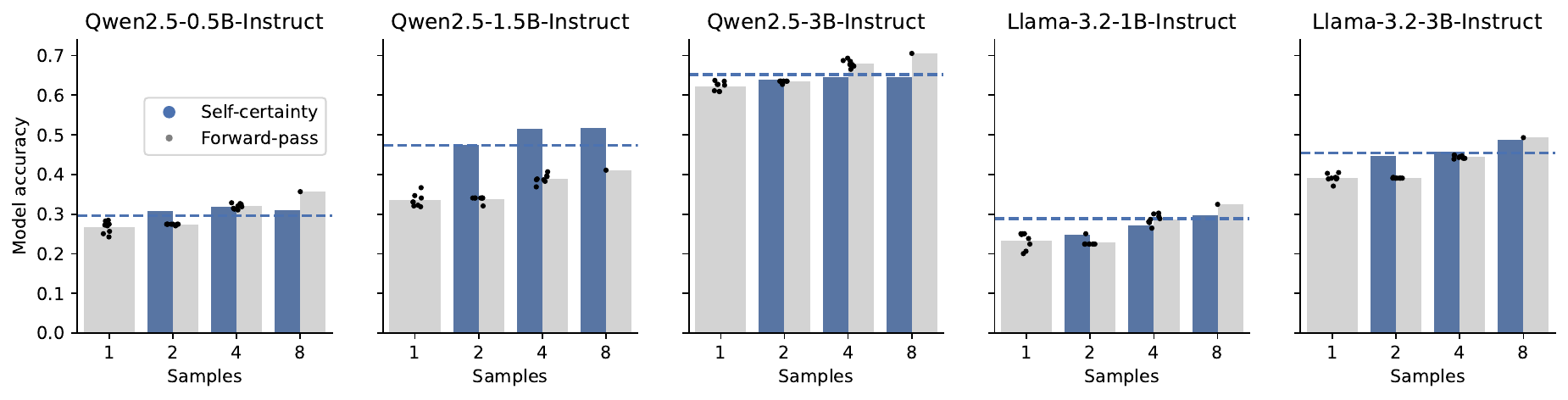}
\caption{\small \textbf{Performance of self-consistency and self-certainty maximisation calling methods across different sizes of the Qwen and Llama family.} Inference-time scaling based on self-certainty maximisation (blue bars) was found to be more efficient than the simple pass (grey bar and black dots), and equalled or outperformed self-consistency (grey bars black dots) with comparable token budgets (samples 2, 4, and 8) in several instances. For the forward pass, each question was evaluated 8 times; the eight dots represent the scores over the full dataset for each generation. The dashed horizontal line indicates the performance of the early stop method, using 16 samples and a maximum of 3 reasoning steps.
}
\label{fig:figure_1}
\end{figure*}
\section{Method} \label{Method}

We provide here some mathematical formalism used in the rest of the paper to describe text generation with an LLM. Queries and questions are provided as text and tokenised into a sequence of tokens denoted $x = (x_1, x_2, ..., x_n)$, $xi \in \mathcal{V}$ where $\mathcal{V}$ is the entire vocabulary. The LLM processes the sequence of tokens $x$ to produce a sequence of logits: $L_x = (\ell_1, \ell_2, ..., \ell_n)$, $\ell_i \in \mathbb{R}^V$ where $V = |\mathcal{V}|$ is the vocabulary size. Each logit vector $\ell_i$ represents the model’s prediction for the $i$-th token, conditional on preceding tokens. For each position $i$, applying the softmax function to $\ell_i$ yields a probability distribution over the vocabulary: $p(\cdot | x_1, ..., x_{i-1}) \in [0, 1]^V$. 

For the last input token $x_n$, the logits $\ell_n$ produce the probability distribution: $p(\cdot | x) \in [0, 1]^V$, representing the likelihood of each token in $V$ being the next token after $x$. After sampling an output sequence $y = (y_1, y_2, ..., y_m)$, the probability distribution for generating the $i$-th token $y_i$, conditional on input $x$ and prior outputs $y_{<i} = (y_1, ..., y_{i-1})$, is: $p(\cdot|x, y_{<i}) \in [0, 1]^V$. This distribution reflects the model’s belief about the next token given the prompt and generated sequence so far.

We define the model's self-certainty $C_i$ regarding the generation of the next $i$-th token as the \textbf{Kullback-Leibler (KL) Divergence} between the output distribution at the token level and a uniform distribution $U$ representing maximum randomness or uncertainty. Following \citep{Kang:2025}, this quantity is given by:

\begin{equation}
\begin{aligned}
C_i &= D_{\mathrm{KL}}(U \,\|\, p(\cdot|x, y_{<i})) \\
    &= \sum_{j=1}^{V} \frac{1}{V}log(\frac{\frac{1}{V}}{p(j|x, y_{<i})}) \\
    &= -\frac{1}{V} \sum_{j=1}^{V} log(V \cdot p(j|x, y_{<i})) \\
\end{aligned}
\end{equation}

We can further generalise and define the self-certainty $C$ at the sentence level as the average of token self-certainty:

\begin{equation} \label{eq_self_certainty}
\begin{aligned}
C &= -\frac{1}{nV}  \sum_{i=1}^{n}\sum_{j=1}^{V} log(V \cdot p(j|x, y_{<i})) \\
\end{aligned}
\end{equation}

Our proposed method, therefore, seeks to sample a set of $k$ reasoning step proposals and select $y^*$ that maximise the self-certainty, such as:

\begin{equation} \label{eq_max_self_certainty}
\begin{aligned}
y^* &= \arg\max_{y_i \in \{y_1, \dots, y_k\}} C(y_i) 
\end{aligned}
\end{equation}

where $y_1, \dots, y_n \sim LLM(x)$. \\

\subsection{Inference-time scaling via self-certainty maximization}

We introduce an inference-time reasoning strategy that selects reasoning steps by maximizing self-certainty, a model-internal confidence signal defined at the level of reasoning steps. A reasoning step (or thought) is defined as the sequence of tokens generated between two predefined reasoning delimiters (Appendix \ref{reasoning_tokens}). This segmentation yields coherent semantic units (e.g., sub-derivations in mathematical reasoning) and enables step-level control during generation.

Self-certainty is quantified as the average token-wise Kullback–Leibler divergence between the model's predictive distribution and the uniform distribution over the vocabulary \citep{Kang:2025, Zhao:2025}. Intuitively, higher self-certainty corresponds to more peaked predictive distributions, indicating stronger internal commitment to a particular continuation.

At inference time, generation proceeds iteratively. At each reasoning step, the model samples $k$ candidate continuations ($k \in {2,4,8}$). Each candidate is scored by its average self-certainty, therefore normalizing by length to ensure comparability across steps. The candidate with the highest score is selected and appended to the context. This process repeats until a valid answer format is produced or a maximum number of reasoning steps is reached (fixed to 40 in all experiments).

This approach has four key properties:
(i) it operates online and does not require pooling complete trajectories;
(ii) it can be applied early in generation, enabling early stopping and reduced token usage;
(iii) it relies exclusively on model-internal signals, avoiding external judges or auxiliary models;
(iv) as evidenced by the results presented below, it improves performance with a limited number of samples while applying to open-ended questions.This has multiple advantaged it for instance makes it compatible with structured generation \citep{willard2023efficient}.

\begin{figure*}[htbp]
\includegraphics[width=\linewidth]{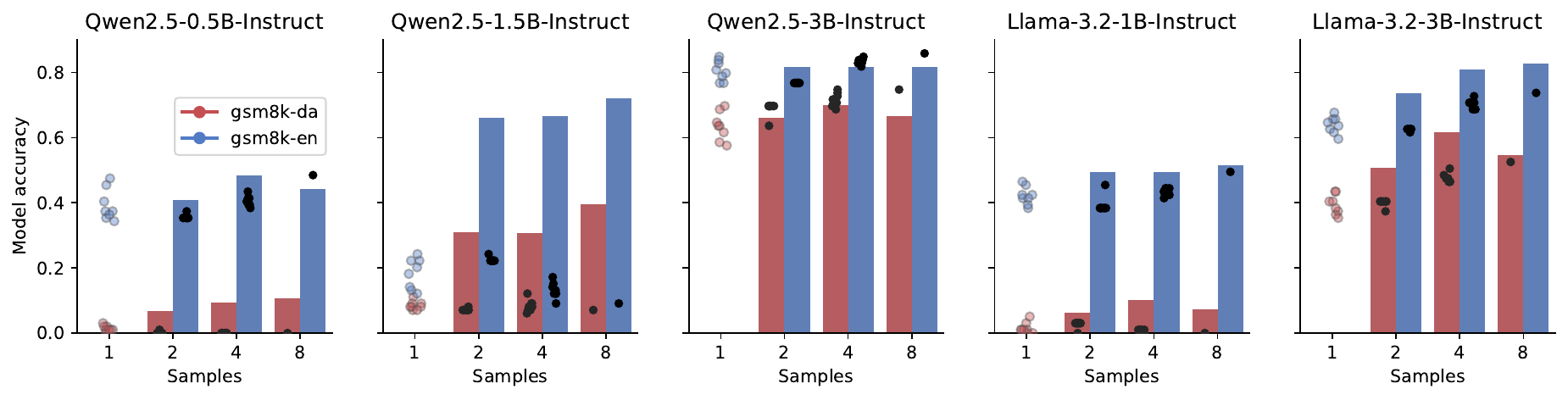}
\caption{\small \textbf{Performance of self-consistency and self-certainty maximisation for English and Danish versions of GSM8K questions, across different sizes of the Qwen and Llama family.} Inference-time scaling methods based on self-certainty maximisation were found more efficient than a single pass (red and blue dots), both with English-based questions from the GSM8k dataset (blue bars), and the same questions translated into Danish (red bars). In a majority of cases, the performances were also higher than self-consistency (black dots) with comparable token budgets (samples 2, 4, and 8).
}
\label{fig:figure_2}
\end{figure*}

\subsection{Evaluation protocol}

We compare self-certainty maximization against inference-time baselines that similarly avoid external supervision. We used one-pass forward greedy decoding as a baseline and self-consistency with majority voting with a comparable number of samples (2, 4, or 8) as a reference. Experiments were conducted on two benchmarks:

500 high-difficulty competition-level problems from the MATH500 dataset \citep{Lightman:2023}.
100 samples of the GSM8K dataset \citep{gsm8k} as samples by \citet{mirzadeh2024gsm} in both their original English form, and translated, verified and localized for Danish\footnote{Datasets is available at \url{https://github.com/centre-for-humanities-computing/m-gsm-symbolic/}}.

For each problem, we evaluated the following inference strategies:

\begin{enumerate}
  \item \textbf{Greedy forward pass}: a single greedy decoding run, serving as the baseline. Each problem is generated eight times to estimate variability.
  \item \textbf{Self-consistency}: Multiple independent reasoning trajectories (2, 4, or 8) are sampled and aggregated via majority voting over final answers \citep{Wang:2022}.
  \item \textbf{Self-certainty maximization (our)}: At each reasoning step, $k \in {2,4,8}$ candidate steps are sampled, and the one maximizing average self-certainty is selected. Generation terminates upon producing a valid answer or reaching 40 reasoning steps.
\end{enumerate}

English prompts follow \citet{Puri:2025} (Appendix~\ref{Eng_Prompt_template}); Danish prompts use a direct translation of the same template (Appendix~\ref{Dan_Prompt_template}).

\subsection{Models}

We evaluate two families of instruction-tuned language models: the \texttt{Qwen2.5-Instruct} model family \citep{qwen:2024} across three sizes: 0.5, 1.5, and 3B parameters and the \texttt{Llama-3.2-Instruct} \citep{llama} across two sizes, 1 and 3B parameters.

These models have been well-suited to explore the efficiency of our inference time scaling methods, and have been reported to increase performance under various approaches \citep{Puri:2025, Agarwal:2025, Zhao:2025}.

\subsection{Analysis of self-certainty dynamics}

We were interested in studying the dynamics of self-certainty during reasoning generation, both under the standard one-pass method and following our sample maximisation strategy. Because we wanted to understand when in the reasoning process the maximisation of self-certainty is likely to exert the greatest influence, we examined the evolution of the average self-certainty across the reasoning steps. We computed this quantity for the forward pass greedy generation, as well as the self-certainty gain that was observed when using sampling. This quantity reflects the improvement gained in selecting the step among the available $k$ samples.

\begin{figure*}[htbp]
\includegraphics[width=\linewidth]{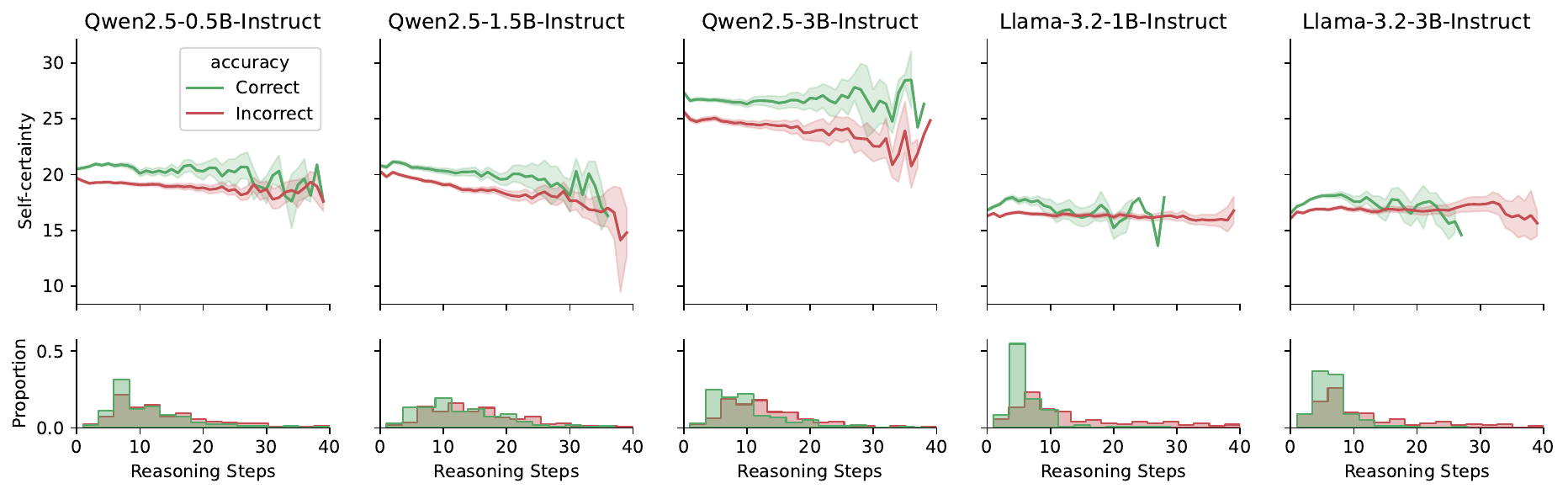}
\caption{\small \textbf{Evolution of self-certainty across reasoning steps for Qwen and Llama models.} Correct trajectories exhibit consistently higher self-certainty than incorrect ones from the earliest reasoning steps. Incorrect trajectories are more frequently associated with long chains of thought.}
\label{fig:figure_3}
\end{figure*}

\subsection{Strategic allocation of inference budget in the reasoning chain}

As a follow-up study, we were interested in varying strategies of computation allocation to reflect the dynamics of uncertainty resolution throughout reasoning. Early steps of reasoning are likely to be critical in LLM as well as relating to high-level control \citep{Griffiths2019} and planning strategies \citep{Lieder2017}. Therefore, we wanted to test whether allocating more budget to these and allocating no budget after $k$ steps would retain or even improve performance.

First, we tested Qwen-3 B's performance on a subset of 100 questions from MATH500. At each step, the model sampled 16 candidates and was allowed to do it for 0, 2, 3, 6, or 40 (all) reasoning steps. Then, we used the results from this experiment to test the accuracy of an optimal sampling procedure by sampling 16 candidates during the first 3 steps only. We tested all the models included in the main benchmarks and reported the results in the same figure.
\section{Results}

In this section, we evaluate self-certainty maximization as an inference-time scaling mechanism for reasoning. We first compare it against greedy decoding and self-consistency under matched token budgets (Section \ref{result_1}). We then assess cross-lingual generalization to a low-resource language (Danish) in Section \ref{result_2}. Finally we analyze the dynamics of self-certainty along reasoning trajectories in Section \ref{result_3} and \ref{result_4}, and study an improved budget allocation strategy that exploits these dynamics in Section \ref{result_5}.

\subsection{Self-certainty maximization improves reasoning accuracy}\label{result_1}

We evaluated self-certainty maximization on MATH500 \citep{Lightman:2023} using Qwen (0.5B, 1.5B, 3B) and Llama (1B, 3B) models (Figure~\ref{fig:figure_1}). The English version of the prompts followed \citep{Puri:2025} and is defined in Appendix \ref{Eng_Prompt_template}. All models were requested to solve the problems using Chain-Of-Thoughts prompting \citep{Wei:2022}. Greedy one-pass decoding (max 1500 tokens) served as the baseline. Self-consistency \citep{Wang:2022} was evaluated under comparable token budgets (i.e., 2, 4 or 8 parallel generations).

Self-certainty maximization proceeds by sampling $n \in {2,4,8}$ candidate reasoning steps at each iteration (up to 40 steps). Candidates are scored using average self-certainty (Eq.~\ref{eq_self_certainty}), controlling for variable step length, and the highest-scoring step is selected and appended to the input for the next generation. For each sequence, we fixed the minimum number of tokens to 5 and the maximum to 300. Generation stopped when a reasoning token was encountered (the complete list of reasoning tokens is defined in Appendix \ref{reasoning_tokens}), and the list of proposals was returned. 

Across model families and sizes, self-certainty maximization consistently matched or outperformed greedy decoding and self-consistency under equivalent budgets (Figure \ref{fig:figure_1}). Performances tended to improve with more tokens, but notably two samples were already enough to observe it. These results indicate that inference-time performance gains do not require large sample counts or external judges, and suggest that smaller models might possess sufficient latent information but struggle with reliable retrieval during greedy decoding.

\subsection{Self-certainty maximisation generalizes across languages}\label{result_2}

To assess robustness beyond English, we evaluated the method on 100 GSM8K problems translated into Danish. The Danish version of the prompt is reported in Appendix \ref{Dan_Prompt_template}. While baseline performance dropped substantially under Danish prompts, self-certainty maximization yielded proportional gains comparable to those observed in English \ref{result_1}. In some cases (e.g., Qwen-1.5B), accuracy improved by up to $4\times$ relative to greedy decoding. This suggests that self-certainty operates as a language-agnostic inference signal, mitigating performance degradation in low-resource or non-English settings.

\begin{figure}[htbp]
\centering
\includegraphics[width=\linewidth, keepaspectratio, height=0.3\textheight]{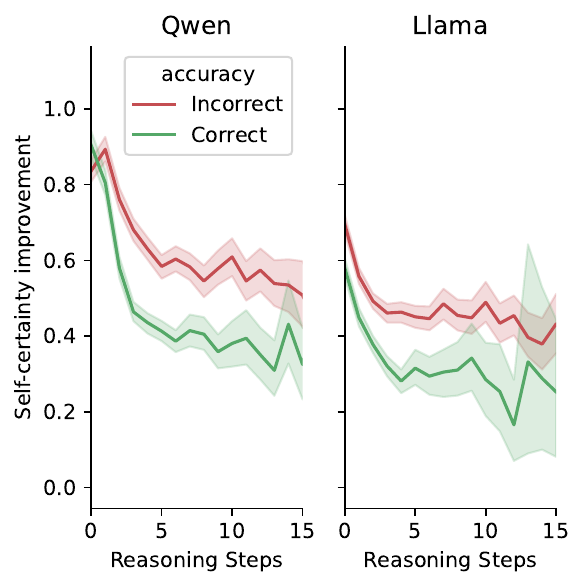}
\caption{\small \textbf{Self-certainty gain across reasoning steps.} Correct trajectories show smaller incremental gains after initialization, indicating early uncertainty resolution.}
\label{fig:figure_4}
\end{figure}

\subsection{Valid trajectories exhibit higher self-certainty throughout reasoning path}\label{result_3}

The raw performance results reported in \ref{result_1} and \ref{result_2} provided encouraging evidence that reasoning performance can be strengthened on small and medium models without relying on costly inference time-scaling methods, either from an external judge or large sample generation. To understand the source of these gains, we analyzed self-certainty dynamics in standard forward generations. For each model, we collected eight greedy rollouts per problem and grouped trajectories by final correctness (we used these different generations to estimate performance uncertainties in Figure \ref{fig:figure_1} and \ref{fig:figure_2}). Self-certainty was computed post hoc at the reasoning-step level.

Across all models (Figure~\ref{fig:figure_3}), correct trajectories exhibited higher self-certainty from the earliest steps. This gap emerged within the first $\sim$20 reasoning steps. Reasoning trajectories that terminated early, with a peak between 5 and 10 reasoning steps, were also more likely to be valid, whereas incorrect trajectories frequently exhausted the maximum reasoning length while exhibiting steadily decreasing self-certainty. These results indicate that signals predictive of eventual correctness are present early in the reasoning process, motivating early stopping or selective exploration strategies.

\subsection{Valid reasoning resolves uncertainty early}\label{result_4}

To directly link self-certainty optimization to correctness, we analyzed trajectories generated via self-certainty maximization. At each step, we measured the self-certainty gain achieved by selecting the best proposal over alternatives given by $\sum_{i=1}^{k}y^* - y_i$ as the self-certainty gain for choosing that step versus another. 

As shown in Figure~\ref{fig:figure_4}, self-certainty gains decrease monotonically along the reasoning path, reflecting reduced ambiguity over time. Crucially, correct trajectories exhibit significantly smaller gains after the first few steps, despite starting at similar levels as incorrect ones. This suggests that successful reasoning rapidly commits to a stable plan, whereas incorrect trajectories continue exploring competing hypotheses. These findings support a view of reasoning as a planning process in which early high-level decisions strongly constrain downstream steps.

\subsection{Strategic allocation of sampling budget}\label{result_5}
 
\begin{figure}[htbp]
\centering
\includegraphics[width=\linewidth, keepaspectratio, height=0.3\textheight]{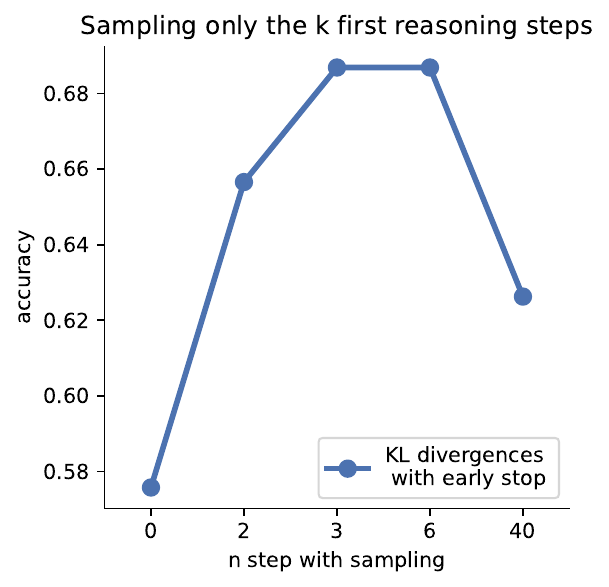}
\caption{\small \textbf{Early self-certainty optimization yields maximal gains.} We tested Qwen-3B against 100 selected questions from MATH500. Sampling is restricted to the first $k$ reasoning steps ($k \in {1,2,5,40}$). Limiting uncertainty minimization to the early steps (i.e., 1 or 2) explains most of the performance increase observed, while optimizing along the entire path decrease gain over time.}
\label{fig:figure_5}
\end{figure}

Motivated by the early concentration of uncertainty resolution (see Figure \ref{fig:figure_4}), we evaluated selective sampling strategies. Using Qwen-3B on 100 MATH500 problems, we fixed the sampling budget to 16 candidates per step but limited sampling to the first $k$ reasoning steps.

Performance followed an inverted U-shape (Figure~\ref{fig:figure_5}). Sampling only during the first 1–5 steps achieved peak accuracy, while sampling at every step degraded performance. This suggests over-optimization, where excessive confidence-based selection leads to brittle or degenerate reasoning paths. These results highlight the importance of when uncertainty is minimized, and point toward adaptive strategies that dynamically allocate inference-time computation.

\subsection{Token budget}\label{result_6}

\begin{figure}[htbp]
\includegraphics[width=\linewidth]{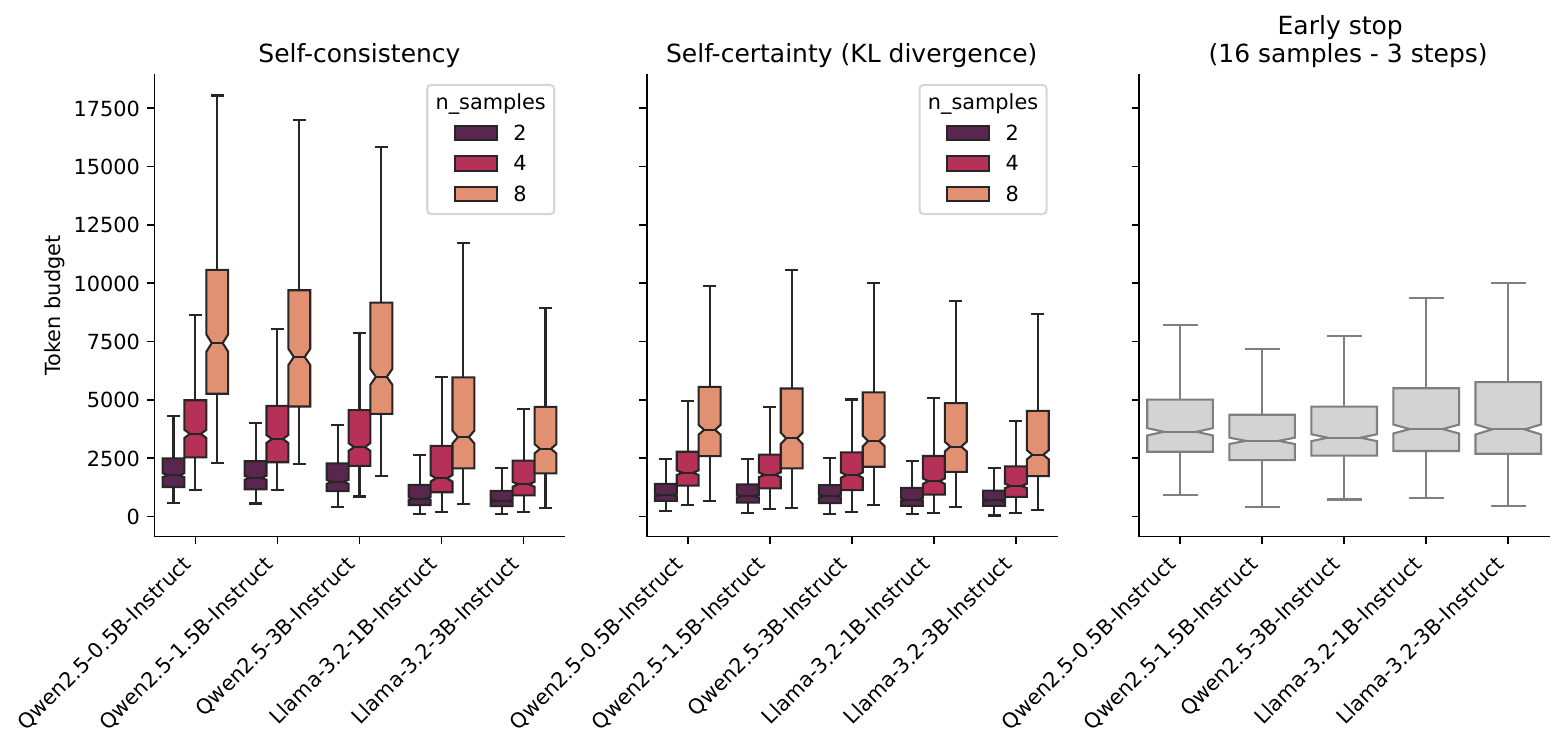}
\caption{\small \textbf{Token usage for the different inference time scaling methods.}}
\label{fig:figure_6}
\end{figure}

\FloatBarrier
\section{Conclusion}

In this work, we introduced an inference-time reasoning strategy based on self-certainty maximization to improve the reasoning performance of small language models. By operating at the level of reasoning steps rather than individual tokens, our approach avoids premature pruning driven by low-probability early tokens and enables principled selection among competing intermediate thoughts. Crucially, the method relies exclusively on model-internal signals and requires neither auxiliary reward models nor additional training, keeping inference-time overhead limited.

We validated this approach against standard inference-time baselines and showed consistent gains across model sizes and with restricted sampling budgets. Beyond performance improvements, our analysis reveals that early reasoning steps play a decisive role in determining final correctness, and that the model’s internal confidence signals are already predictive at these early stages. This finding suggests that inference-time scaling strategies that concentrate additional computation on later phases of generation may yield diminishing returns, and that reallocating compute toward early reasoning decisions is a more effective path for improving reasoning accuracy.

\section*{Acknowledgments}

N.L. is supported by Danish Foundation Models (4378-00001B) and the European Union (101178170). K.N. is supported by Danish Foundation Models (4378-00001B), The Carlsberg Foundation (CF23-1583), European Union (101178170), Danish Research Foundation (DNRF193), the Aage and Johanne Louis-Hansens Foundation and Augustinus Foundation.

\section*{Code and data availability}

To support reproducibility, the code for this work will be released upon acceptance.

\section*{Large language model usage}

We used large language models to assist with proofreading and improving the clarity of the manuscript. The authors are solely responsible for the content and any remaining errors.
w

\bibliography{ref}

\appendix

\section{Appendix}

\subsection{English Prompt template} \label{Eng_Prompt_template}

\begin{tcolorbox}[colback=gray!10, colframe=gray!80, boxrule=0.5pt, arc=2pt, left=6pt, right=6pt, top=4pt, bottom=4pt, title= English Evaluation System Prompt]

Solve the following math problem efficiently and clearly:

\hspace*{2em}- For simple problems (2 steps or fewer): \\
\hspace*{2em}Provide a concise solution with minimal explanation.

\vspace{1em}

\hspace*{2em}- For complex problems (3 steps or more): \\
\hspace*{2em}Use this step-by-step format: \\

\vspace{1em}

\hspace*{2em}\#\# Step 1: [Concise description] \\
\hspace*{2em}[Brief explanation and calculations]

\vspace{1em}

\hspace*{2em}\#\# Step 2: [Concise description] \\
\hspace*{2em}[Brief explanation and calculations]

\vspace{1em}

\hspace*{2em}...

\vspace{1em}

Regardless of the approach, always conclude with:

\vspace{1em}

Therefore, the final answer is: \textbackslash boxed\{answer\}. I hope it is correct.

\vspace{1em}

Where [answer] is just the final number or expression that solves the problem.

\end{tcolorbox}

\subsection{Danish Prompt template} \label{Dan_Prompt_template}

\begin{tcolorbox}[colback=gray!10, colframe=gray!80, boxrule=0.5pt, arc=2pt, left=6pt, right=6pt, top=4pt, bottom=4pt, title= Danish Evaluation System Prompt]

Løs følgende matematiske problem effektivt og tydeligt: \\

\hspace*{2em}- For nemme problemer (2 trin eller færre): \\
\hspace*{2em}Giv en præcis løsning med minimal forklaring. \\
    
\hspace*{2em}- For komplekse problemer (3 trin eller mere): \\
\hspace*{2em}Brug dette trinvise format:

\vspace{1em}

\hspace*{2em}\#\# Trin 1: [Koncis beskrivelse] \\
\hspace*{2em}[Kort forklaring og beregninger]

\vspace{1em}

\hspace*{2em}\#\# Trin 2: [Koncis beskrivelse] \\
\hspace*{2em}[Kort forklaring og beregninger]

\vspace{1em}

\hspace*{2em}...

\vspace{1em}

Uanset fremgangsmåde, afslut altid med: \\

Derfor er det endelige svar: \textbackslash boxed\{answer\}. Jeg håber, det er korrekt.

\vspace{1em}

Hvor [answer] er blot det endelige tal eller udtryk, der løser problemet.

\end{tcolorbox}

\subsection{Reasoning tokens} \label{reasoning_tokens}

The following tokens were defined as reasoning token, and would therefore stop the generation and return the thought as step proposal:

\begin{center}
\begin{verbatim}
<think>
</think>
.\n\n
\n\n
:\n\n
]\n\n
)\n\n
.\n\n
).\n\n
):\n\n
\end{verbatim}
\end{center}

We defined all double line breaks as reasoning tokens to encourage short reasoning steps and avoid long trajectories.

\end{document}